\pdfoutput=1
\documentclass[11pt]{article}

\usepackage{naacl2021}

\usepackage{times}
\usepackage{latexsym}
\usepackage{amsmath,graphicx}
\usepackage{multirow}
\usepackage{booktabs}
\usepackage{caption}
\usepackage{subcaption}
\makeatletter
\def\blfootnote{\xdef\@thefnmark{}\@footnotetext}
\makeatother
\usepackage[T1]{fontenc}
\usepackage[utf8]{inputenc}

\usepackage{microtype}

\title{Pushing the performances of ASR models on English and Spanish accents}

\author{Pooja Chitkara*, Morgane Riviere*, Jade Copet, Frank Zhang, Yatharth Saraf \\ Meta AI, USA}

\begin{document}
%Datasets names
\newcommand{\es}[1]{$\text{es}_{\text{#1}}$}
\newcommand{\en}[1]{$\text{en}_{\text{#1}}$}

\newcommand{\esar}{\es{ar}} %en_ar
\newcommand{\esgt}{\es{gt}} %en_gt
\newcommand{\eses}{\es{es}} %en_es
\newcommand{\espe}{\es{pe}} %en_pe
\newcommand{\esve}{\es{ve}} %en_ve
\newcommand{\eslat}{\es{lat}} %en_latam
\newcommand{\esmx}{\es{mx}} %es_mx
\newcommand{\escl}{\es{cl}} %es_mx
\newcommand{\escar}{\es{car}} %es_caribe
\newcommand{\escam}{\es{cam}} %es_central_america
\newcommand{\esap}{\es{ap}} %es_ando_pacifico
\newcommand{\vphundred}{VP-100k} % VoxPopuli 100k
\newcommand{\indomaines}{ID-100k}
\newcommand{\commonvoices}{c.v.}
\newcommand{\userdata}{user data}
\newcommand{\avp}{avp}
\newcommand{\nisp}{nisp}
\newcommand{\allsstar}{allsstar}

\newcommand{\usagespanish}{usage-es}
\newcommand{\usageenglish}{usage-en}
%Models
\newcommand{\transformerbig}{transformer}
\newcommand{\wavvec}{wav2vec}
\newcommand{\emformer}{emf.}

%Coefficients
\newcommand{\lambdaasr}{\lambda_{asr}}
\newcommand{\lambdaacc}{\lambda_{acc}}

%New commands
\newcommand*{\morgane}{\textcolor{red}{@morgane}~\textcolor{red}}
\newcommand*{\pooja}{\textcolor{blue}{@pooja}~\textcolor{blue}}
\newcommand*{\tmp}{\textcolor{purple}}

\maketitle
\begin{abstract}
Speech to text models tend to be trained and evaluated against a single target accent.
This is especially true for English for which native speakers from the United States became the main benchmark.
In this work, we are going to show how two simple methods: pre-trained embeddings and auxiliary classification losses can improve the performance of ASR systems.
We are looking for upgrades as universal as possible and therefore we will explore their impact on several models architectures and several languages.
\end{abstract}
\blfootnote{* equal contribution}
\section{Introduction}

Speech to text models tend to be trained on data representing a limited number of accents.
In English for example, US English has become the norm while there are many other English speaking countries on the globe.
Moreover, non-native English speakers make the vast majority of the English speaking world and Automatic Speech recognition (ASR) models tend to recognize them poorly ~\cite{riviere2021asr4real}.

This observation also holds for other languages.
For example Spanish is the official languages of 20 sovereign states, totalling more than 440 million people.
Arabic is an official language of 23 countries and is spoken by almost as many people as Spanish.
Mandarin with more than 1 billion native speakers shows a significant variety of accents.

In this work we are going to explore two simple methods to improve the resilience of ASR models against diverse accents: pre-training and auxiliary accent classification loss.
We will focus on two languages: English and Spanish. 
Furthermore, we will also consider different models and architecture and see if these upgrades can apply effectively to all of them.
With our experiments, we will be able to show the following points:
\begin{itemize}
\itemsep -1mm
    % \item We pushed several state of the art models in order to get better performances on specific accents
    \item Auxiliary accent classification has a positive impact on accented data
    \item Pretrained embeddings improve performances even with a large amount of fine-tuning data  and transfer well from one language to another.
    %\item Theses improvement applies also to accents absent from the training data.
    %\item Theses improvements made our models converge faster
\end{itemize}

\section{Related Work}

\textbf{Accented data}. There have been several efforts to improve ASR models when dealing with accented audio.
Benchmark works have been done so far like \cite{koenecke7684} who worked on African American speakers, \cite{riviere2021asr4real} who evaluated several models in very diverse settings.
Moreover, \cite{jain2018improved} and \cite{Yang2018Joint} used multi-task learning to improve the performances of a model on accented English.
\textbf{Multilingual ASR}. On top of that, combining several languages to improve the performance of a model becomes more and more common in Speech recognition.
As far as pre-training is concerned for example, we can quote the work of \cite{wangvoxpopuli}, \cite{bai2021joint} or \cite{conneauxlsr} with strong benchmarks on public datasets.
Besides, \cite{pratap2020massively} recently released a massively multilingual ASR system to improve the coverage of low resource languages in speech recognition.

\section{Approach}

In this paper, we test two simple and standard methods on two languages, English and Spanish, to see if they can reliably improve the resilience of ASR models when facing different accents. 

\subsection{Auxiliary accent classification}

In this approach, we propose a multi-task learning (MTL) architecture  where the encoder of an ASR system jointly learns to classify an accent label per utterance \cite{jain2018improved} \cite{li2021accentrobust}.
We consider both native and non-native accents for English as well as several native accents in Spanish. 
The complete list of our accent labels is available in tables \ref{tab:UsageEnglishAccents}, \ref{tab:UsageSpanishAccents}, and \ref{tab:PublicEnglishAccents} of the appendix.

%We empirically compare the effectiveness of positioning this auxiliary accent classification layer at the lower, middle, and top layers of the encoder. 
%We also experiment with adding linear weights to the auxiliary accent cross-entropy loss and the encoder asr cross-entropy loss before summing them to backpropagate through the shared layers of the encoder. 
%Finally, we demonstrate the usefulness of selectively zero-ing out the loss from the accent predictor for utterances with noisy accent labels. 

\subsection{Pre-training}

%\subsection{Semi-supervised Pretraining}

% Self-training \cite{schuber} is a classical semi-supervised approach to improve the performance of a model using unlabelled data.
% The idea is simple: unlabelled data are annotated using the inference of a teacher model and then we launch a new training stage using both labelled data and these weakly annotated unlabelled data.

% In this paper we are going to see if self-training can be used as a pre-training stage.
% Labelled and unlabelled English data are plentiful and we suspect that we can use a model pre-trained on a big English dataset and transfer it to another language. 
% But can it also make the model more resillient when facing diverse accents ?

%\subsection{Unsupervised pre-training}

Still today, many state of the art ASR models rely on pre-computed speech features like Mel Filterbanks. 
However, we are now seeing increasingly more models replacing them with fully trained embeddings directly put on the raw audio waveform.
This pretraining stage can be done using only unlabelled data, which can be a huge advantage for languages with very few labelled data points available.

Unsupervised pre-training, has shown very good performances both in and out of domain.
For example Wav2vec2.0 \cite{baevski2020wav2vec} trained on 60kh of unlabelled data allows near state of the art results on Librispeech with less than an hour of transcribed speech.
Furthermore, both \cite{conneauxlsr} and \cite{riviere2021asr4real} demonstrated that pretraining was very resilient to both language change and domain transfer. 
Therefore, if embeddings can be successfully pre-trained on a language and fine-tuned on another one, can pre-training help models with diverse accents?

%We will analyze the performances of a Wav2Vec2.0 model when changing the pre-training dataset and how it compares to models trained in a semi-supervised setting.
%Why: pretraining transfers both between languages   and between domain . Can it help with accent resilience ?

\section{Datasets}

We use both public and internal data for our experiments: they are summed up in table \ref{tab:all_data}.

\subsection{Training Data}

Our Spanish dataset contains about 3,000 hours and our English dataset contains about 6,000 hours of audio extracted from de-identified public videos. 
To protect users’ privacy, we cannot make these datasets public. 
Each dataset is augmented using speech perturbation at rates 0.1 and 0.9, and segmented into 10s-long chunks.
As for now, we will refer to our labelled training data as \usageenglish~and \usagespanish.
%We augment each dataset using speed perturbation resulting in 30,000 hours of Spanish and 60,000 hours of English data, and segment them into 10s-long chunks. 
%We randomly sample 10\% of the augmented dataset per language to generate 3000 hours of Spanish training data (referred to throughout as \usagespanish) and 6000 hours of English training data (referred to throughout as \usageenglish).
% If enough time: launch a training on English with public data (VoxPopuli with accent labels + CommonVoices with accent labels + ?)

\subsection{Evaluation data}

\subsubsection{User data}

We evaluate the spanish models against 6 manually annotated accents in \usagespanish. %, of which all accents except \eslat, which represents a mix of Latin American accents, are manually annotated. 
To benchmark \usageenglish, we consider 27 native and non-native accents. %, of which all accents except \en{us} are manually annotated and \en{gb} contains a mix of accents from England, Scotland, and Ireland. 
The detailed distribution of these datasets is available in Tables \ref{tab:UsageEnglishAccents} and \ref{tab:UsageSpanishAccents} of the Appendix.
%The distribution of duration per accent to benchmark \usagespanish is represented in Table \ref{tab:\usagespanishAccents} and to benchmark \usageenglish is shown in Table \ref{tab:\usageenglishAccents}.

\begin{table}[h!]
    \centering
    \begin{tabular}{c|c|c|c|c}
    \toprule
      Language & Split & Source & $N_a$ & $S$ (h)\\
      \midrule
      \multirow{5}{*}{English}  & Train &\usageenglish & 27& 6k\\
       \cmidrule{2-5}
      & \multirow{ 4}{*}{Test} &\userdata & 27& 489\\
      &  &allsstar & 22& 21\\
      &  &nisp & 5 & 29\\
      &  &\avp &23 & 143\\
      \midrule
      \midrule
      \multirow{3}{*}{Spanish}  & Train &\usagespanish & 6& 3k\\
      \cmidrule{2-5}
      &\multirow{2}{*}{Test}  & user data & 6& 149\\
      &&\commonvoices &4 & 3.7\\
    \bottomrule
    \end{tabular}
    \caption{
      \textbf{An overview of our datasets} $N_a$ being the number of accents represented in the dataset and $S$ its size. \userdata~cannot be released publicly. In the Common Voices dataset (\commonvoices), we considered only the sequences with an accent label. 
    \label{tab:all_data}}
\vspace{-4mm}
\end{table}

\subsubsection{Public benchmark}

On top of our private benchmark we also consider an evaluation on public data.
For English, we will use the accented datasets presented in ~\cite{riviere2021asr4real} : ALLSSTAR ~\cite{allsstar}, NISP ~\cite{kalluri2020nisp} and Accented VoxPopuli ~\cite{wangvoxpopuli}.
Besides, for Spanish we will use the dev and test sets of Mozilla's Common Voices (v5) ~\cite{ardila2020common}. 
However, given the small amount of dev and test Common Voices data with an accent label, we had to add all Common Voices validated data to our evaluation dataset.

\section{Models}

To begin, we will do our ablation study on two standard letter-based, acoustic models (AM) used in academia.%: one of them using directly the raw audio waveform and the other relying on MFSC embedding.
Then, we will combine our methods to push the performances of a model optimized for inference on user data. %word-piece based Enformer model using log-mel filters embeddings.

\subsection{Academic models}

We consider letter based AM trained with the Connectionist Temporal Classification loss \cite{ctc}. 
In order to fully apprehend how each of our upgrade impacts the AM, we did not use a Language model at the inference stage.

\textbf{Transformer model}.
The RASR models developed by \cite{likhomanenko2021rethinking} showed impressive performance in English for both different speech settings and accents.
They rely on a targeted data augmentation applied to a large fully labelled training set in order to train model that transfers well to new domains.
We will consider a smaller version of the architecture developed in RASR: our model will be composed of a convolutional frontend with width $7$ and stride $3$, followed by a GLU activation, a sinusoidal positional embedding and finally $24$ 4-heads transformer blocks. 
These blocks have a self-attention dimension of $768$ and an inner feed forward dimension of $3072$.

\textbf{Wav2vec 2.0}. We consider here the implementation in the flashlight codebase of the Wav2vec2.0 method as described in \cite{chai2020cpcwav2vec}. 
The model is an encoder followed by a context network put directly on top of the raw audio waveform.
The encoder contains seven convolutional layers of hidden dimension $512$ followed by a ReLU activation while the context is a $12$ layers transformer network with embeddings of dimension $768$, $8$ heads and feed forward layers of dimension $3072$. 

\subsection{Word piece based Emformer model}

Our baseline model follows an emformer-based \cite{shi2021emformer} RNN-T architecture with auxiliary cross-entropy prediction layers \cite{liu2021improving}. 
We fully describe our architecture in the appendix.
For better performances, we add auxiliary cross-entropy losses at the 5th, 11th, 17th, and 23rd layers of  the Spanish model and 6th, 13th, 20th, and 27th layers of the English one.
We encode transcriptions using a fixed-size vocabulary of word pieces from a Sentence Piece \cite{kudo2018sentencepiece} model trained on these transcripts with a vocabulary size of 2k for Spanish and of 5k for English. 
We apply Specaugmentation \cite{park2019specaugment} on our input data and use adam optimizer with tri-stage learning rate schedule \cite{wu2020streaming} with peak learning rate of 0.0004 trained over 110 epochs.
Finally, at inference time, we use Voice Activity Detection to ignore noisy samples and a decoding threshold of 800ms.

\section{Results}

In our experiments, we consider the global WER on each dataset.
In other words, we divide the total number of errors in the entire dataset and then divide it by its total number of words.

\subsection{Ablation study}
\subsubsection{Accent Classifier}
\addtolength{\tabcolsep}{-1.5pt} 
\begin{table*}[t!]
    \centering
     \begin{tabular}{l|c|cc|ccccc|ccccc}
    \toprule
      \multirow{2}{*}{Model} &\multirow{2}{*}{$N_p$}&\multirow{2}{*}{MTL}& \multirow{2}{*}{PT} &\multicolumn{5}{c}{User data benchmark}  & \multicolumn{5}{c}{Common Voices} \\
        &&&& \esar & \eses & \esgt & \espe & \esve &\esap & \escar & \es{spa} & \esmx & \escam   \\
      \midrule
      \multirow{3}{*}{\wavvec}& \multirow{3}{*}{94M} & No& No& 29.4 & 47.4 & 34.1 & 34.3 & 29.7 & 14.3& 16.7 & 18.4 & 16.7 & 21.5 \\ 
     % &&No& OD & \tmp{14.3} & x.xx & x.xx & x.xx & x.xx & x.xx& x.xx & x.xx & x.xx & x.xx \\ 
      &&No& UE & \textbf{25.0} & \textbf{43.6} & \textbf{30.9} & \textbf{28.6} & \textbf{26.6} & \textbf{11.1}& \textbf{12.9} &\textbf{15.5} & \textbf{13.9} & \textbf{15.0} \\ 
      &&Yes & UE & 27.6 & 45.9 & 32.7 & 32.2 & 29.5 & 12.7 & 15.6 &17.6 & 16.0 & 16.9 \\ 
      \midrule
      %\multicolumn{4}{l}{\textit{Mel-filterbanks based models}} \\
      \midrule
      \multirow{2}{*}{\transformerbig}  &\multirow{2}{*}{256M}&No& No & 31.9 & 45.6 & 43.2 & 36.7 & 46.0 & 12.3 & 15.3 & 15.3 & 15.0& 27.2 \\ 
      && Yes & No &\textbf{28.1} & \textbf{42.4} & \textbf{38.3} & \textbf{29.9} & \textbf{33.1} & \textbf{10.7} & \textbf{13.7} & \textbf{13.7} & \textbf{14.1}& \textbf{20.1} \\ 
      %&& Yes & SE & x.xx& x.xx & x.xx & x.xx & x.xx & x.xx&x.xx & x.xx & x.xx & x.xx & x.xx \\ 
    %   \midrule
    %   \midrule
    %   \multirow{3}{*}{\enformer}&\multirow{3}{*}{??} & No & No& 14.4 & 13.0 & 13.6 & 15.2 & 15.6 &  10.8 & 13.1 & 13.3 & x.xx & 23.0\\ 
    %   && Yes & No & 14.0 & 12.5 & 13.3 & 14.9 & 15.4 & 9.87 & 12.7 & 13.1 & x.xx & \textbf{12.9}\\ 
    %   && Yes & SE & \textbf{13.4} & \textbf{11.9} & \textbf{12.6} & \textbf{14.2} & \textbf{14.5} & \textbf{8.06} & \textbf{10.5} & x.xx & x.xx &18.1 \\ 
     \bottomrule
      \end{tabular}
    \caption{
      \textbf{Average WER on Spanish accents on both an internal and a public benchmark when combining the MTL loss and the pretraining.} All models were trained on \usagespanish. UE stands for Unsupervised pre-training on European languages while $N_p$ refers to the number of parameters of the model.
    \label{tab:combo_es}}
\end{table*}
\addtolength{\tabcolsep}{1.5pt} 
\addtolength{\tabcolsep}{-1.5pt} 
\begin{table*}[t!]
    \centering
     \begin{tabular}{l|cc|ccc|ccc|ccc|ccc}
    \toprule
      \multirow{2}{*}{Model} &\multirow{2}{*}{MTL}& \multirow{2}{*}{PT} &\multicolumn{3}{c}{User Data}  & \multicolumn{3}{c}{NISP} & \multicolumn{3}{c}{AVP} & \multicolumn{3}{c}{ALLSSTAR}\\
      &&& $w_{all}$  & $w_{top}$ & $w_{bot}$& $w_{all}$  & $w_{top}$ & $w_{bot}$& $w_{all}$  & $w_{top}$ & $w_{bot}$& $w_{all}$  & $w_{top}$ & $w_{bot}$\\
      \midrule
      \multirow{3}{*}{\wavvec} &No& No& 24.9  &13.0 & 35.1 & 21.1 & 18.1 & 25.4&  22.1& 19.4 & 31.3 & 20.4 & 13.4 & 27.7\\ 
     % &&No& OD & \tmp{14.3} & x.xx & x.xx & x.xx & x.xx & x.xx& x.xx & x.xx & x.xx & x.xx \\ 
      &No& UE &\textbf{22.7} & \textbf{11.9} & \textbf{31.4} & \textbf{18.7} & \textbf{16.2} & \textbf{22.7}& \textbf{20.6} & \textbf{17.4} & \textbf{27.8} & \textbf{17.8} & \textbf{12.6} & \textbf{26.6} \\ 
      &Yes & UE & 27.3 & 12.6 & 34.3 & 20.7 & 17.6 & 25.0 & 21.2 & 18.4 & 29.3 & 28.4 & 17.1 & 41.2 \\
      \midrule
      %\multicolumn{4}{l}{\textit{Mel-filterbanks based models}} \\
      \midrule
      \multirow{2}{*}{\transformerbig}  &No& No &20.4 & 10.9 & 28.1 & \textbf{19.1} & \textbf{17.0} & \textbf{26.4} & 21.3 & \textbf{18.5} & \textbf{29.2} & 27.7 & 18.3 & 35.7\\ 
      &Yes& No & \textbf{20.5} &\textbf{10.4} & \textbf{28.9}  & 19.5 &17.2 & 27.6 & \textbf{21.2} & \textbf{18.5} & 31.4 & \textbf{21.7} & \textbf{13.5} & \textbf{30.9} \\
      %&& Yes & SE & x.xx& x.xx & x.xx & x.xx & x.xx & x.xx&x.xx & x.xx & x.xx & x.xx & x.xx \\ 
    %   \midrule
    %   \midrule
    %   \multirow{3}{*}{\enformer}&\multirow{3}{*}{??} & No & No& 14.4 & 13.0 & 13.6 & 15.2 & 15.6 &  10.8 & 13.1 & 13.3 & x.xx & 23.0\\ 
    %   && Yes & No & 14.0 & 12.5 & 13.3 & 14.9 & 15.4 & 9.87 & 12.7 & 13.1 & x.xx & \textbf{12.9}\\ 
    %   && Yes & SE & \textbf{13.4} & \textbf{11.9} & \textbf{12.6} & \textbf{14.2} & \textbf{14.5} & \textbf{8.06} & \textbf{10.5} & x.xx & x.xx &18.1 \\ 
     \bottomrule
      \end{tabular}
    \caption{\textbf{Average WER on our English benchmark with academic models.} All models were trained on the \usageenglish~dataset. We consider the global WER $w_{all}$, the best WER on a given accent $w_{top}$ and the worst one $w_{bot}$.
    \label{tab:combo_en}}
\vspace{-2mm}
\end{table*}
\addtolength{\tabcolsep}{1.5pt} 

\textbf{Transformer model}.
We try to put the auxiliary accent classifier at different layers of the model, and we find the best results when adding the auxiliary classification loss to the last transformer layer of the architecture, with a weight $\lambdaacc=0.01$.
As shown in table ~\ref{tab:combo_es} for Spanish, the use of the MTL loss flattens the performances discrepancies between each accents and leads to a model more robust and more accurate on average. 

\textbf{Wav2vec 2.0}.
For this experiment, we use a Wav2vec model pretrained on user data (see section below). 
We try out several possible positions and weights for the auxiliary accent classification loss, but unfortunately, without seeing any performance improvement.
As shown in tables ~\ref{tab:combo_es} and ~\ref{tab:combo_en}, the MTL loss systematically leads to a slight increase of the average WER on each language and accent considered.

\subsubsection{Pre-training}

We consider two settings for the Wav2vec architecture: in the first one, we do not perform any kind of pre-training, on the second one we pre-train our own model using 25,000 hours of unlabelled user data in various European languages.

As we can see in tables \ref{tab:combo_es} and \ref{tab:combo_en}, even with a large amount of labelled data, unsupervised pre-training still gives a significant edge to Wav2vec models.
Even better, the Wav2vec model pre-trained in a fully unsupervised fashion competes with the much bigger Transformer model.

\subsection{Pushing the 'real-life' model further}

\addtolength{\tabcolsep}{-1.5pt} 
\begin{table}[t!]
    \centering
     \begin{tabular}{l|ccccc}
    \toprule
        Model & \esar & \eses & \esgt & \espe & \esve \\
      \midrule
       \emformer & 14.4 & 13.0 & 13.6 & 15.2 & 15.6 \\ 
      \emformer + mtl & 14.0 & 12.5 & 13.3 & 14.9 & 15.4\\ 
      \emformer + mtl + ss & \textbf{13.4} & \textbf{11.9} & \textbf{12.6} & \textbf{14.2} & \textbf{14.5} \\ 
      \midrule
      \midrule
        Model & \en{nat} & \en{lat} & \en{eur} & \en{afr} & \en{eaa} \\
        \midrule
       \emformer & 15.8 & 21.4 & 15.5 & 26.2 & 24.3 \\ 
      \emformer + mtl & \textbf{15.5} & \textbf{21.2} & \textbf{15.2} & \textbf{25.9} & \textbf{21.3} \\ 
      \bottomrule
      \end{tabular}
    \caption{
    \label{tab:combo_enformer} \textbf{Average WER on user data using model optimized for user data}. We consider the following speakers : native English speakers \en{nat}, Latin Americans \en{lat}, Europeans \en{eur}, Africans \en{afr} and East Asians \en{eaa}. ss stands for semi-supervised pre-training. }
\vspace{-4mm}
\end{table}
\addtolength{\tabcolsep}{1.5pt} 

We start by applying the accent classification loss to the emformer model.
%\textbf{Enformer model}. We started to experiment with the 'accent prediction' approach applied to the Emformer model. 
%We used the following settings: 'all' indicates that we backpropagate the accent classification cross-entropy loss for all accents in UsageSpanish while 'clean' indicates that we set the accent classification loss to zero for utterances which do not have gold annotated labels ie for utterances with label \eslat. 
%Since English does not have noisy labels that cover multiple accent types, we only keep the 'all' setting with UsageEnglish. 
%Parameters $\lambdaacc$ and $\lambdaasr$ indicate the linear weights of auxiliary losses from the accent classifier and the auxiliary asr layers respectively. 
%The total auxiliary loss can be expressed as the weighted sum of losses from the aux accent classifier and the aux asr layers: 
%We call $\lambdaasr$ the weight of the auxiliary asr loss of the model, then the total auxiliary loss becomes:
%\[L_{aux} = \lambdaacc* L_{acc} + \lambdaasr * L_{asr}\]
%Finally 'low', 'middle', and 'top' indicate the position of the aux classifier layer in the emformer as the 8th, 14th, and the 20th layers for Spanish, and at the 9th, 16th, and 23rd layers for English respectively. 
%We observe that the best setting uses clean labels, places the accent classifier in the middle, and allocates a small weight of 0.1 to the aux accent loss and 0.9 to the aux asr loss. 
We observe that the best setting places the accent classifier at an intermediate layer of the emformer model (the 14th one for Spanish and the 16th one for English) and with $\lambdaacc=0.1$ and $\lambdaasr=0.9$, where $\lambdaasr$ is the weight of the auxiliary asr loss of the model. 
The detailed results of these experiments can be found in tables \ref{tab:mtlUsageSpanish}, \ref{tab:mtlUsageEnglishAll}, and \ref{tab:mtlUsageEnglishAllDetailed} in the appendix.
As shown in table \ref{tab:combo_enformer} this loss leads to a 2.4\% improvement of the WER for Spanish and a 1.52\% one for English.
%The detailed results on our English benchark is available in table \ref{tab:mtlUsageEnglishAllDetailed} of the appendix.
%The average accent word error rate improves by 2.4\% for Spanish and by 1.52\% for English (see table \ref{tab:combo_enformer}). 
%We show the word error rates per accent across different hyperparameter settings in Table \ref{tab:mtlUsageSpanish} for Spanish, in the appendix. 
%We compare average word error rates for English per setting in Table \ref{tab:mtlUsageEnglishAll}, in the appendix. 
Then, we combine this auxiliary loss with semi-supervised pretraining
% Then, we combine this auxiliary loss with some pretraining.
% We couldn't transfer the unsupervised wav2vec pretraining to this architecture therefore we fell back to semi-supervised pretraining.

Self-training is a classical semi-supervised approach to improve the performance of a model using unlabelled data.
The idea is simple: unlabelled data are annotated using the inference of a teacher model and then we launch a new training stage using both labelled data and these weakly annotated unlabelled data.
We can leverage a very large amount of English data labelled by high quality teacher models, but we can not do the same in Spanish. 
But we already know that self-training has a positive impact when done within the same language \cite{kahn2020selftrain}, so can it transfer from one language to another?
To observe that, we consider a model with architecture identical to our baseline English Emformer, trained using 14k hours of gold transcribed audio, similar to \usageenglish, and 2 million hours of semi-supervised social media videos as a pre-trained seed to our Emformer model for Spanish with our best-performing accent classifier settings.% of 'clean, mid, $\lambdaacc=0.1$, $\lambdaasr=0.9$'. 
%We initialize the linear embedding weights and the 24 emformer encoder layers with a stride of 8, using the English teacher model weights. 
We train all parameters, including the pre-trained weights, of the Spanish Emformer model with accent classifier. 
This leads us to a relative average improvement of 7.29\% in average accent word error rate compared to the baseline. 
%Table \ref{tab:accentPlusPretrain} in the appendix, shows the change in wer per accent using our proposed approach of combining accent classifier and pre-training.

%\subsubsection{Unseen accents}

%Proposition: For english only (not a lot of data on es and ar), remove 1 easy, 1 medium and 1 hard accent from the training set and see how it transfers.
%\input{tables/unseen_accents}

\section{Conclusion}

We have observed that for both English and Spanish, the use of an auxiliary loss on accent classification could bring a moderate improvement to the performances of two of the architectures we considered.
However, it didn't work well with wav2vec.
Our paper also brings another proof that pre-training, both supervised and unsupervised transfers well across languages.
Besides, it stays effective even with a big fine-tuning dataset.
%Pre-training also still brings a significant upgrades to the performances even when a large amount of labelled data is available for fine-tuning.
Therefore, we think that the large amount of both public and private high quality English data could be leveraged to build resilient ASR systems in other languages. 
Finally, both of these techniques work for all the accents considered in our study: in Spanish they work for speakers from both Spain and Latin America while on English they also improve the performances of models on non-native speakers.

To go further, we think this work could be expanded to other languages, especially non-European ones like Arabic or Mandarin.

\newpage
\bibliography{anthology,custom}
\bibliographystyle{acl_natbib}
\newpage
\appendix

\section{Enformer model}
We start by extracting 80-dimensional log-mel filters per 10ms of audio, downsampled with stride 6 for Spanish and stride 8 for English, and project it into a 512-dimensional space with a linear layer. 
The output is fed into a multi-layer emformer (24 layers for Spanish and 28 layers for English), where each layer contains 8 multi-head attention heads and a feed-forward of dimension 2048 with a dropout of 0.1. 
The output is projected into 1024 dimensions with a linear layer followed by a layer normalization. 
In the predictor, the tokens are first represented by 256-dimensional embeddings which are passed through two LSTM layers of 512 dimensions, followed by a linear projection into 1024-dimensions. 
The joiner first has a layer of tanh activation followed by a linear projection to the target number of word pieces. 
\section{Datasets}

In tables ~\ref{tab:UsageSpanishAccents} and ~\ref{tab:UsageEnglishAccents} you will find the detailed composition of our internal benchmark based on user data.
\begin{table}[h!]
    \centering
    \begin{tabular}{l|c|c|c}
    \toprule
      Country& Label & Group &  Duration (h)\\
      \midrule
      New Zealand &  \en{nz} & \multirow{8}{*}{\en{nat}} &17 \\
      Scotland &\en{sq} && 20 \\
      Canada&\en{ca} && 19 \\
      UK &\en{gb} && 23 \\
      England &\en{en} && 24 \\
      Ireland &\en{ie} && 21 \\
      Australia & \en{au} && 24 \\ 
      US &\en{us} && 23 \\
      \midrule
      Columbia &\en{co} & \multirow{4}{*}{\en{lat}} &  19 \\
      Brazil & \en{br} & & 19 \\ 
      Mexico &\en{mx} & &16 \\ 
      Argentina &\en{ar} & &10 \\ 
      \midrule
      Germany & \en{de}& \multirow{4}{*}{\en{eur}} & 15 \\ 
      Italy & \en{it} &&10 \\
      France &\en{fr} && 26 \\
      Russia &\en{ru} && 10 \\
      \midrule
      Taiwan &\en{tw} & \multirow{8}{*}{\en{eaa}} &11 \\
      Philippines &\en{ph} && 19 \\
      Myanmar & \en{mm} &&  17 \\ 
      Pakistan & \en{pk} &&  20 \\
      Korea &\en{kr} && 21 \\ 
      Vietnam &\en{vn} && 20 \\
      Bangladesh &\en{bd} && 10 \\
      India &\en{in} && 24 \\
      \midrule
      Egypt & \en{eg} &  \multirow{3}{*}{\en{afr}} &10 \\
      Nigeria &\en{ng} && 21 \\
      South Africa &\en{za} && 20 \\
      \midrule
      \multicolumn{3}{l}{\textbf{TOTAL}} & 489 \\
    \bottomrule
    \end{tabular}
    \caption{
      \textbf{Duration (hours) per accent to benchmark models trained with \usageenglish}. We consider five different groups of speakers : native speakers \en{nat}, Latin American Speakers \en{lat}, European Speakers \en{eur}, East Asian Speakers \en{eaa} and African speakers \en{afr}.
    \label{tab:UsageEnglishAccents}}. 
\end{table}
\addtolength{\tabcolsep}{-1.5pt} 
\begin{table}[h!]
    \centering
    \begin{tabular}{l|l|c|c}
    \toprule
      \multirow{2}{*}{Dataset} & Country& \multirow{2}{*}{Label} & Duration\\
      & / Region & & (h) \\
      \midrule
      \multirow{6}{*}{\usagespanish}& Argentina &\esar  & 19 \\
      &Guatemala & \esgt  & 24 \\
      &Spain& \eses & 23 \\ 
      &Peru&\espe & 24 \\ 
      &Venezuela& \esve  & 24 \\ 
      &Latin America & \eslat & 35 \\
      \midrule
      \multirow{5}{*}{\commonvoices} & Andean countries& \esap & 0.6\\
      & Caribbean & \escar & 0.4\\
      & Spain & \es{spa} & 1.5\\
      & Mexico & \esmx & 1  \\
      & Central America & \escam & 0.2\\
      \midrule
      \multicolumn{3}{l}{\textbf{TOTAL}} &  152.7\\
    \bottomrule
    \end{tabular}
    \caption{
      \textbf{Duration per accent of our Spanish evaluation data}. In the Common Voices dataset (\commonvoices), accents are grouped by large geographic units rather than by country.
    \label{tab:UsageSpanishAccents}}
\end{table}
\addtolength{\tabcolsep}{1.5pt} 
\begin{table}[h!]
    \centering
    \begin{tabular}{l|c}
    \toprule
      Dataset & Native languages \\
      \midrule
      \multirow{2}{*}{\nisp} & Hindi, Kannada, Malayalam, \\
       &Tamil, Telugu  \\
      \midrule 
      \multirow{8}{*}{\allsstar} & English, Mandarin, Cantonese, \\
      & Farsi, French, German \\
      & Gishu, Greek,  Gujarati \\
      & Hebrew, Hindi, Icelandic \\
      & Indonesian, Japanese, Korean \\
      & Malay, Portuguese, Russian \\
      & Spanish, Thai, Turkish \\
      &Vietnamese\\
      \midrule
      \multirow{8}{*}{\avp} & Bulgarian, Czech, Danish\\
      & German, Greek, Spanish \\
      & Estonian, Finnish, French \\
      & English (Ireland), Croatian \\
      & Hungarian, Italian, Lithuanian \\
      & Latvian, Maltese, Dutch \\
      & Polish, Portuguese, Romanian \\
      & Slovak, Slovene, Swedish \\
    \bottomrule
    \end{tabular}
    \caption{
      \textbf{Native languages of the speakers represented in our public English test sets}.
    \label{tab:PublicEnglishAccents}}
\end{table}

\section{Detailed performances of the Enformer Model}

We used the following settings: 'all' indicates that we backpropagate the accent classification cross-entropy loss for all accents in UsageSpanish while 'clean' indicates that we set the accent classification loss to zero for utterances which do not have gold annotated labels ie for utterances with label \eslat. 
Since English does not have noisy labels that cover multiple accent types, we only keep the 'all' setting with UsageEnglish. 
Finally 'low', 'middle', and 'top' indicate the position of the aux classifier layer in the emformer as the 8th, 14th, and the 20th layers for Spanish, and at the 9th, 16th, and 23rd layers for English respectively. 
\begin{table*}[h!]
    \centering
    \begin{tabular}{cc|cc|cccccc|cc}
    \toprule
      Position &labels& $\lambdaacc$ & $\lambdaasr$ & \esar &\eses &\esgt &\espe &\esve & \eslat & avg & $\Delta$ (\%)\\
      \midrule
      baseline&& -& -& $14.36$ & $13.03$ & $13.55$ & $15.15$ & $15.57$ & $13.52$ & $14.19$ & -\\
      \midrule
      \multirow{3}{*}{low}& all& $0.1$& $0.9$ & $14.14$ & $12.59$ & $13.32$ & $14.91$ & $15.39$ & $13.27$ & $13.94$ &$-1.76$ \\
      &clean& $0.1$& $0.9$ & $14.05$ & $12.74$ & $13.30$ & $14.84$ & $15.40$ & $13.24$ & $13.93$ &$-1.83$\\
      &clean& $0.2$& $0.8$ & $14.17$ & $12.76$ & $13.35$ & $14.88$ & $15.45$ & $13.39$ & $14.00$&$-2.40$\\
      \midrule
      \multirow{2}{*}{mid}&clean& $0.1$&$0.9$ & $13.98$ & $12.54$ & $13.34$ & $14.88$ & $15.41$ & $13.27$ & $13.90$ & $-1.34$\\
      &clean& $0.2$& $0.8$ & $14.08$ & $12.80$ & $13.37$ & $14.99$ & $15.40$ & $13.36$ & $14.00$ &$-1.34$\\
      \midrule
      \multirow{2}{*}{top}&all& $0.1$& $0.9$ & $14.11$ & $12.57$ & $13.32$ & $14.98$ & $15.38$ & $13.39$ & $13.96$ &$-1.62$\\
      &clean& $0.1$& $0.9$ & $14.04$ & $12.56$ & $13.30$ & $15.00$ & $15.42$ & $13.28$ & $13.93$ & $-1.83$\\

    \bottomrule
    \end{tabular}
    \caption{
      \textbf{WER with auxiliary accent prediction task on Emformer model for Spanish}. The MTL classifier is tested at different layers of the model: low being the 8th one, 'middle' the 14th and top the last enformer layer. In the 'all' setting, we backpropagate the MTL loss for all samples, while in the 'clean' setting we ignore noisy accent labels.
    \label{tab:mtlUsageSpanish}}
\end{table*}
\begin{table}[h!]
    \centering
    \begin{tabular}{c|c|c|c|c}
    \toprule
      Position & $\lambdaacc$ & $\lambdaasr$ & wer & \% change\\
      \midrule
      baseline & - & -& 14.19& -\\
      \midrule
      \multirow{3}{*}{low}  & $1.0$ & $1.0$ & 20.58 & +3.46\%\\
      &$0.5$& $1.0$ & 20.23 & +1.71\%\\
      & $0.1$ & $0.9$ & 19.67 & -1.08\%\\
      \midrule
      \multirow{3}{*}{mid}& $1.0$& $1.0$ & 20.59 & +3.57\%\\
      &$0.5$& $1.0$ & 20.11 & +1.11\%\\
      &$0.1$&$0.9$ & 19.59 & \textbf{-1.52\%}\\
      \midrule
      \multirow{3}{*}{top}& $1.0$& $1.0$ & 21.36 & +7.41\%\\
      & $0.5$&$1.0$ & 20.57 & +3.43\%\\
      & $0.1$& $0.9$ & 19.74 & -1.04\%\\
    \bottomrule
    \end{tabular}
    \caption{
      \textbf{Average WER with auxiliary accent prediction task on Emformer model for English}. The baseline was trained without accent classification.
    \label{tab:mtlUsageEnglishAll}}
\end{table}
\begin{table}[h!]
    \centering
    \begin{tabular}{c|c|c|c}
    \toprule
      Accent & Baseline & With MTL & Rel\% change\\
      \midrule
      $en_{us}$ & 15.26 & 15.17 & -0.55\% \\
      $en_{co}$ & 22.14 & 21.96 & -0.81\% \\
      $en_{eg}$ & 25.20 & 24.93 & -1.08\% \\
      $en_{mm}$ & 28.37 & 28.42 & +0.16\% \\
      $en_{nz}$ & 13.98 & 13.54 & -3.09\% \\
      $en_{pk}$ & 27.32 & 27.26 & -0.22\% \\
      $en_{br}$ & 19.92 & 19.45 & -2.37\% \\
      $en_{de}$ & 11.36 & 11.13 & -2.04\% \\
      $en_{ru}$ & 15.82 & 15.63 & -1.22\% \\
      $en_{tw}$ & 22.95 & 22.42 & -2.29\% \\
      $en_{ph}$ & 27.49 & 26.52 & -3.53\% \\
      $en_{mx}$ & 21.65 & 21.54 & -0.48\% \\
      $en_{sq}$ & 16.68 & 16.28 & -2.41\% \\
      $en_{it}$ & 16.14 & 15.89 & -1.54\% \\
      $en_{fr}$ & 18.49 & 17.86 & -3.37\% \\
      $en_{kr}$ & 12.32 & 12.13 & -1.57\% \\
      $en_{vn}$ & 32.21 & 31.82 & -1.23\% \\
      $en_{ca}$ & 10.75 & 10.66 & -0.84\% \\
      $en_{gb}$ & 20.89 & 20.87 & -0.11\% \\
      $en_{ar}$ & 22.49 & 22.04 & -2.01\% \\
      $en_{bd}$ & 21.39 & 21.23 & -0.71\% \\
      $en_{en}$ & 16.29 & 15.89 & -2.43\% \\
      $en_{in}$ & 21.91 & 21.77 & -0.66\% \\
      $en_{ng}$ & 27.09 & 26.91 & -0.65\% \\
      $en_{ie}$ & 15.32 & 14.93 & -2.53\% \\
      $en_{za}$ & 19.09 & 18.41 & -3.56\% \\
      $en_{au}$ & 14.45 & 14.14 & -2.14\% \\
    \bottomrule
    \end{tabular}
    \caption{
      \textbf{Average WER per accent with auxiliary accent prediction task on Emformer model for English}. The baseline was trained without accent classification.
    \label{tab:mtlUsageEnglishAllDetailed}}
\end{table}

\label{sec:appendix}
% I have moved the text to another file: the text is left untouched.

\end{document}